\documentclass[10pt,twocolumn,letterpaper]{article}

\usepackage[pagenumbers]{cvpr} 

\usepackage[dvipsnames]{xcolor}
\definecolor{cvprblue}{rgb}{0.21,0.49,0.74}
\usepackage[pagebackref,breaklinks,colorlinks,citecolor=cvprblue]{hyperref}
\usepackage{booktabs}
\usepackage{multirow}
\usepackage{algorithm}
\usepackage{algorithmic}
\usepackage{makecell}
\usepackage[accsupp]{axessibility}
\usepackage{amsmath}

\begin{document}

\title{Delving into Decision-based Black-box Attacks on Semantic Segmentation}

\author{Zhaoyu Chen$^{1,2}$\footnotemark[1]$\ $$\quad$
        Zhengyang Shan$^{1}$\footnotemark[1]$\ $ $\quad$
        Jingwen Chang$^{1}\quad$
        Kaixun Jiang$^{1}$\\
        Dingkang Yang$^{1}\quad$
        Yiting Cheng$^{2}\quad$ 
        Wenqiang Zhang$^{1,2}$ \\ 
        $^1$ Shanghai Engineering Research Center of AI \& Robotics, Academy for Engineering \& Technology, \\ Fudan University $\quad$
        $^2$ Shanghai Key Lab of Intelligent Information Processing, \\ School of Computer Science, Fudan University \\
        {\tt\small{zhaoyuchen20@fudan.edu.cn}}
}

\maketitle

\renewcommand{\thefootnote}{\fnsymbol{footnote}}
\footnotetext[1]{indicates equal contributions.} 

\begin{abstract}
Semantic segmentation is a fundamental visual task that finds extensive deployment in applications with security-sensitive considerations. Nonetheless, recent work illustrates the adversarial vulnerability of semantic segmentation models to white-box attacks. However, its adversarial robustness against black-box attacks has not been fully explored. In this paper, we present the first exploration of black-box decision-based attacks on semantic segmentation. First, we analyze the challenges that semantic segmentation brings to decision-based attacks through the case study. Then, to address these challenges, we first propose a decision-based attack on semantic segmentation, called \textbf{D}iscrete \textbf{L}inear \textbf{A}ttack (\textbf{DLA}). Based on random search and proxy index, we utilize the discrete linear noises for perturbation exploration and calibration to achieve efficient attack efficiency. We conduct adversarial robustness evaluation on 5 models from Cityscapes and ADE20K under 8 attacks. DLA shows its formidable power on Cityscapes by dramatically reducing PSPNet's mIoU from an impressive 77.83\% to a mere 2.14\% with just 50 queries.
\end{abstract}

\section{Introduction}
\label{sec:intro}

Deep neural networks (DNNs) have made unprecedented advancements and are extensively employed in various fundamental vision tasks, such as semantic segmentation~\cite{fcn,pspnet,deeplabv3} and video object segmentation~\cite{hong2021adaptive,hong2022lvos,guo2022adaptive}. However, recent studies have revealed the susceptibility of DNNs to adversarial examples~\cite{fgsm,chen2022towards,chen2022shape,chen2023contentbased} by adding specially designed small perturbations to the input that are imperceptible to humans. 
The emergence of adversarial examples has prompted researchers to focus on the security of underlying visual tasks and seek inspiration for the development of robust DNNs through the exploration of adversarial examples.

Semantic segmentation is a primary visual task for pixel-level classification. Despite its extensive utilization in real-world safety-critical applications like autonomous driving and medical image segmentation, it remains susceptible to adversarial examples. Recently, the emergence of Segment Anything Model (SAM)~\cite{sam} has attracted people's attention to segmentation models and inspired exploration of their robustness. 
However, there are few adversarial attacks on semantic segmentation~\cite{segpgd}, and they focus more on white-box attacks. White-box attacks require access to all information about the model (e.g., gradients and network architecture), which is challenging and often unattainable in real-world scenarios. 
Consequently, black-box attacks offer a more effective means to explore the adversarial robustness of semantic segmentation models in real-world scenarios.

In this paper, we explore for the first time black-box attacks on semantic segmentation in the decision-based setting. The decision-based setting represents the most formidable challenge among black-box attacks, as it restricts access solely to the output category provided by the target model, without any information regarding probabilities or confidences. 
Nevertheless, the efficacy of decision-based attacks on semantic segmentation remains severely constrained by the inherent characteristics of pixel-level classification, as evidenced by the following observations: 
\textbf{i)~Inconsistent Optimization Goal.} In image classification, decision-based attacks often reduce the magnitude of perturbations under the premise of misclassification. However, in semantic segmentation, the larger the perturbation amplitude, the lower the metric, and it is difficult to constrain the perturbation to the $l_p$ norm. \textbf{ii)~Perturbation Interaction.} Perturbations from different iterations interfere with each other, so a pixel is classified incorrectly in this iteration but may be classified correctly under perturbation in the next iteration, which leads to optimization difficulties. \textbf{iii) Complex Parameter Space}. Attacking semantic segmentation is a multi-constraint optimization problem, wherein the complexity of the parameter space imposes limitations on attack efficiency. In practice, it becomes imperative to employ an efficient decision-based black-box attack to assess the adversarial robustness of semantic segmentation. Therefore, the proposed attack must exhibit both high attack efficiency and reliable attack performance.

To tackle the aforementioned challenges, we first propose the decision-based attack on semantic segmentation, termed \textbf{D}iscrete \textbf{L}inear \textbf{A}ttack (\textbf{DLA}). DLA employs a random search framework to effectively generate adversarial examples from clean images, utilizing a proxy index to guide the optimization process. Specifically, we optimize the adversarial examples by leveraging the changes in the proxy index corresponding to the image. Additionally, we alleviate the challenges identified in Section~\ref{sec:analysis} by proposing discrete linear noises for updating the adversarial perturbation. For interference between perturbations, we find that locally adding noises has a good attack effect, but the added colorful patches are easily perceived. Therefore, we introduce linear noises and update the perturbation by adding horizontal or vertical linear noises to the image. To further compress the parameter space, we convert the complex continuous parameter space into a discrete parameter space and bisect the discrete noise from the extreme point of the $l_\infty$-norm ball. 
The overall process can be divided into two parts: perturbation exploration and perturbation calibration. In perturbation exploration, DLA adds discrete linear noises to the input to obtain a better initialization. In the perturbation calibration, DLA adaptively flips the perturbation direction of some regions according to the proxy index, updates and calibrates the perturbation.
We evaluate the adversarial robustness of semantic segmentation models based on convolutional neural networks (FCN~\cite{fcn}, PSPNet~\cite{pspnet}, DeepLabv3~\cite{deeplabv3}) and transformer (SegFormer~\cite{segformer} and Maskformer~\cite{maskformer}) on Cityscapes~\cite{cityscapes} and ADE20K~\cite{ade20k}. Extensive experiments demonstrate that DLA achieves state-of-the-art attack efficiency and performance on semantic segmentation.
Our main contributions and experiments are as follows:

\begin{itemize}
    \item We first explore the adversarial robustness of existing semantic segmentation models based on decision-based black-box attacks, including CNN-based and transformer-based models.
    \item We analyze and summarize the challenges of decision-based attacks on semantic segmentation.
    \item We first propose the decision-based attack on semantic segmentation, called Discrete Linear Attack (DLA), which applies discrete linear noises to perturbation exploration and perturbation calibration.
    \item Extensive experiments show the adversarial vulnerability of existing semantic segmentation models. On Cityscapes, DLA can reduce PSPNet’s mIoU from 77.83\% to 2.14\% within 50 queries.
\end{itemize}

\section{Related Work}
\label{sec:relatedwork}

\textbf{Semantic Segmentation.}~Semantic segmentation is a visual task of pixel-level classification.~Currently, DNN-based methods have become the dominant way of semantic segmentation since the seminal work of Fully Convolutional Networks (FCNs)~\cite{fcn}.~The subsequent model focuses on aggregating long-range dependencies in the final feature~map: DeepLabv3~\cite{deeplabv3} uses atrous convolutions with various atrous rates and PSPNet~\cite{pspnet} applies pooling technology with different kernel sizes.~The subsequent work began to introduce transformers~\cite{transformer} to model context: SegFormer~\cite{segformer} replaces convolutional backbones with Vision Transformers (ViT)~\cite{vit} that capture long-range context starting from the very first layer. MaskFormer~\cite{maskformer} introduces the mask classification and employs a Transformer decoder to compute the class and mask prediction.

\noindent\textbf{Black-box Adversarial Attack. }
In this paper, we primarily concentrate on query-based black-box attacks, where it is assumed that attackers have limited access to the target network and can only make queries to obtain the network's outputs (confidences or labels) for specific inputs~\cite{qeba,devopatch}. The former are called score-based attacks, while the latter are decision-based attacks. Generally speaking, score-based attacks have higher attack efficiency on image classification. For decision-based attacks on semantic segmentation, we define the model output as the label of each pixel. Considering that the mIoU of semantic segmentation is a continuous value calculated based on the label of each pixel, we choose score-based attacks on image classification as the competitors in this paper. Most score-based attacks on image classification estimate the approximate gradient through zeroth-order optimizations~\cite{nes}. Bandits~\cite{bandits} further introduce the gradient prior information and bandits to accelerate~\cite{nes}. Then, \citeauthor{zosignpgd}~\cite{zosignpgd} introduce the zeroth-order setup to sign-based stochastic gradient descent (SignSGD)~\cite{signpgd} and propose ZO-SignSGD~\cite{zosignpgd}. Then, SignHunter~\cite{signhunter} exploits the separability property of the directional derivative and improves the query efficiency.
Recently, methods based on random search have been proposed and have better query efficiency. SimBA~\cite{simba} randomly samples a vector from a predefined orthonormal basis to images. Square Attack~\cite{square} selects localized square-shaped updates at random positions to update perturbations. 
Compared with previous work, DLA analyzes the challenges of semantic segmentation and implements query-efficient attacks based on discrete linear noise.

\noindent\textbf{Adversarial Attack on Semantic Segmentation. }
Compared to image classification, there are few adversarial attacks on semantic segmentation.~\cite{segrobustness} and \cite{dag} are the first to study the adversarial robustness of semantic segmentation and illustrate its vulnerability through extensive experiments. Indirect Local Attack~\cite{indirect} reveals the adversarial vulnerability of semantic segmentation models due to long-range context.
SegPGD~\cite{segpgd} improves white-box attacks from the perspective of loss functions and can better evaluate and boost segmentation robustness. ALMA prox~\cite{proximal} produces adversarial perturbations with much smaller $l_\infty$ norms with a proximal splitting. The aforementioned attacks primarily prioritize enhancing the strength of white-box attacks on semantic segmentation, while allocating comparatively less emphasis on the adversarial robustness of query-based black-box attacks. Consequently, as a complementary approach, we undertake the pioneering exploration of adversarial robustness within the highly challenging decision-based setting.

\section{Method}
\label{sec:method}

\begin{figure}[t]
    \centering
    \includegraphics[width=1\linewidth]{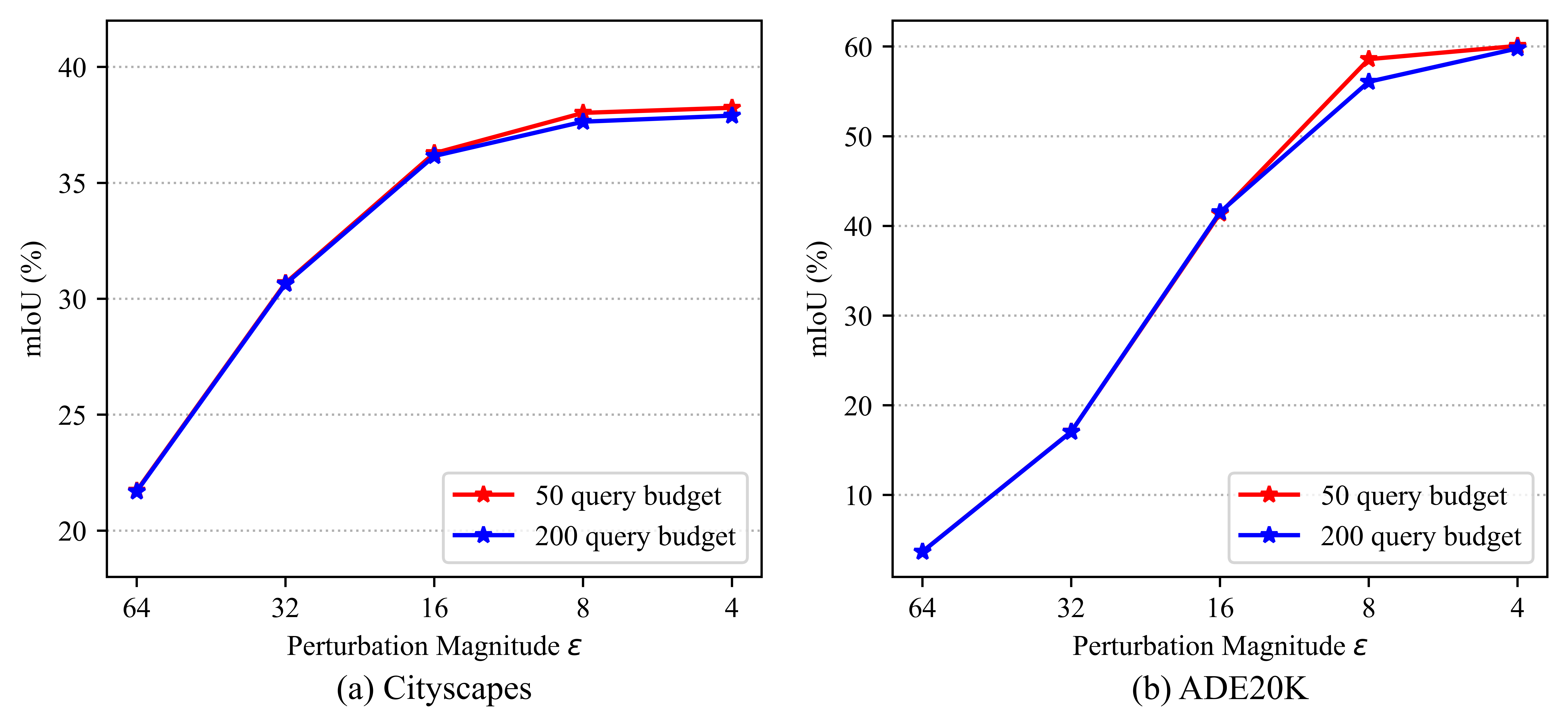}
    \caption{Based on Random attack, we give the changes in mIoU under various perturbation magnitudes. If we add a very large perturbation, this can make the mIoU very small. However, when reducing the perturbation magnitude, the mIoU increases, which makes the optimization goal and attack direction inconsistent.}
    \label{fig:eps}
    \vspace{-12pt}
\end{figure}

\subsection{Preliminaries}
In semantic segmentation, given the semantic segmentation model $f(\cdot)$, the clean image is $x \in [0,1]^{C\times H\times W}$ and the corresponding labels are $y_i \in \{1,...,K\}^{d}$ ($d=HW$ and $i=1,...,d$), where $C$ is the number of channels, $H$ and $W$ are the height and width of the image, and $K$ is the number of semantic classes. We denote the adversarial example $x_{adv}=x+\delta$, where $\delta^{C\times H\times W}$ is the adversarial perturbation and it satisfies $||\delta||_\infty \leq \epsilon$. Because the attack is the decision-based setting, we denote the model output as the per-pixel predicted labels $\hat{y}=f(x)\in \{1,...,K\}^d$. We hope that the adversarial example can make all pixels misclassified as much as possible, so the optimization goal can be expressed as:
\begin{equation}
\label{equ:opt}
\begin{aligned}
        & \underset{\delta}{\arg\max} \sum \mathsf{1}(f(x+\delta)_i \neq y_i), \\ 
        & \mathrm{s.t.}\   ||\delta||_\infty \leq \epsilon \ \mathrm{and}\ i=1,...,d,
\end{aligned}
\end{equation}
where $\mathsf{1}$ is the indicator function. When the condition is met, it is recorded as 1, otherwise it is 0.

\subsection{Attack Analysis}
\label{sec:analysis}
Decision-based attacks on image classification have been extensively and intensively studied~\cite{qeba}, however, semantic segmentation has not been fully explored. Semantic segmentation is a pixel-level classification, which is far more difficult to attack than image classification, because attacking image classification is a single-constraint optimization, while every pixel in semantic segmentation must be classified, which results in attacking semantic segmentation as a multi-constraint optimization. Consequently, decision-based attacks on semantic segmentation encounter substantial challenges, often leading to optimization convergence towards local optima, as illustrated below.

\begin{figure}[t]
    \centering
    \includegraphics[width=1\linewidth]{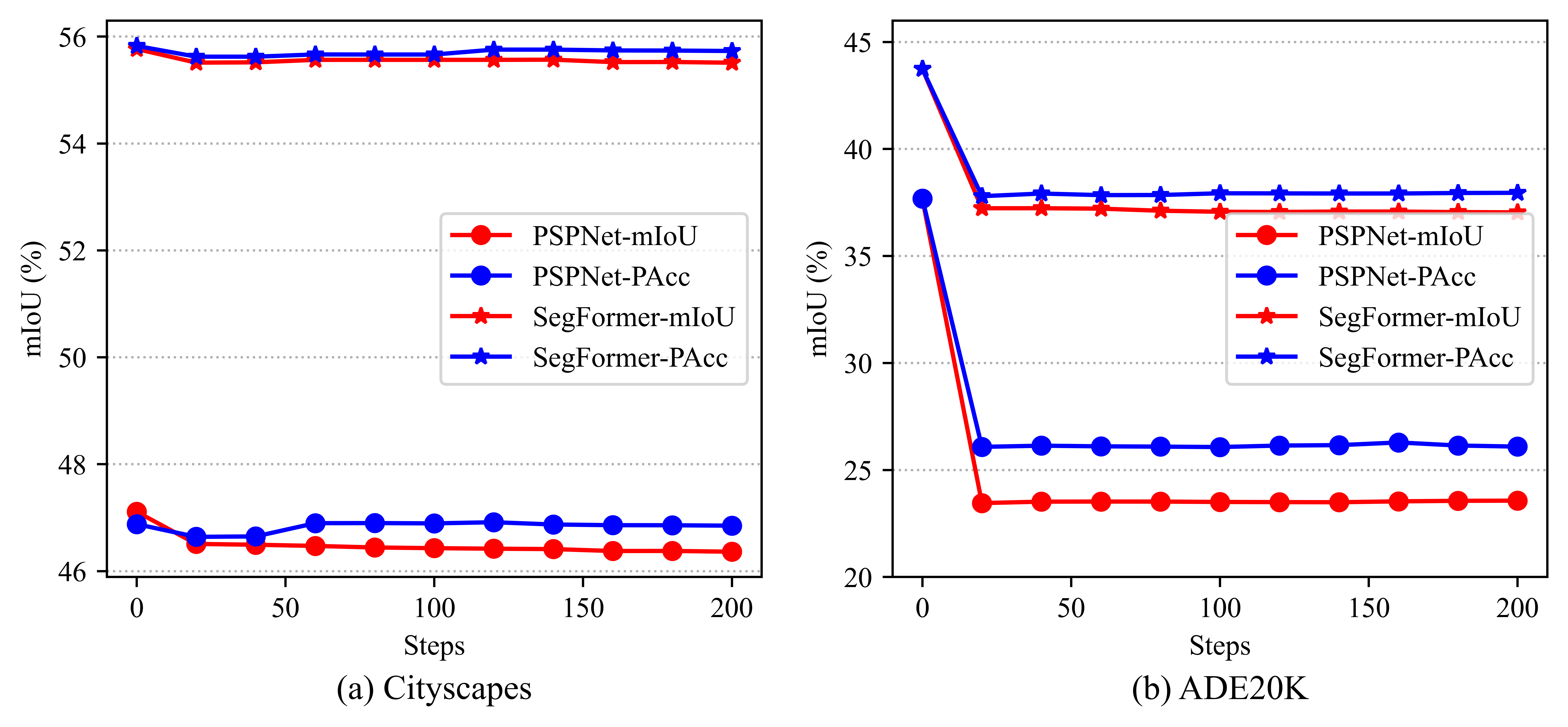}
    \caption{Random attack with different proxy indexes. Our design focuses on optimizing the adversarial perturbation by initiating from clean images and iteratively updating the example based on the observed changes in the proxy index.}
    \label{fig:proxyindex}
\end{figure}

\begin{figure}[t]
    \centering
    \includegraphics[width=1\linewidth]{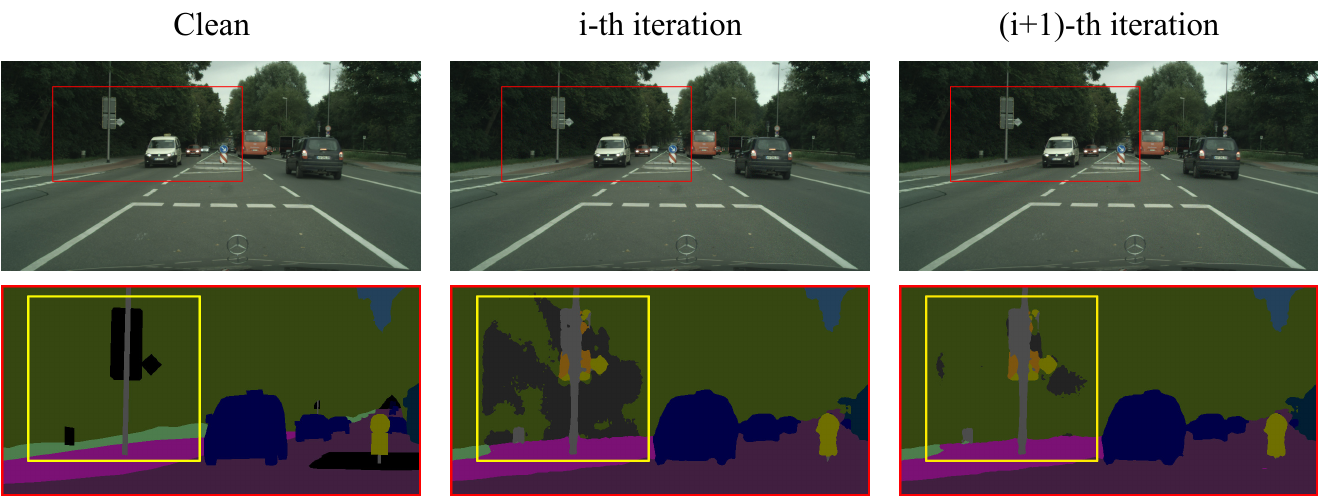}
    \caption{When facing black-box attacks on semantic segmentation, the update of perturbations causes the attacked pixels to revert to their original categories, resulting in optimization difficulties.}
    \label{fig:recover}
\end{figure}

\begin{figure*}[t]
    \centering
    \includegraphics[width=1\linewidth]{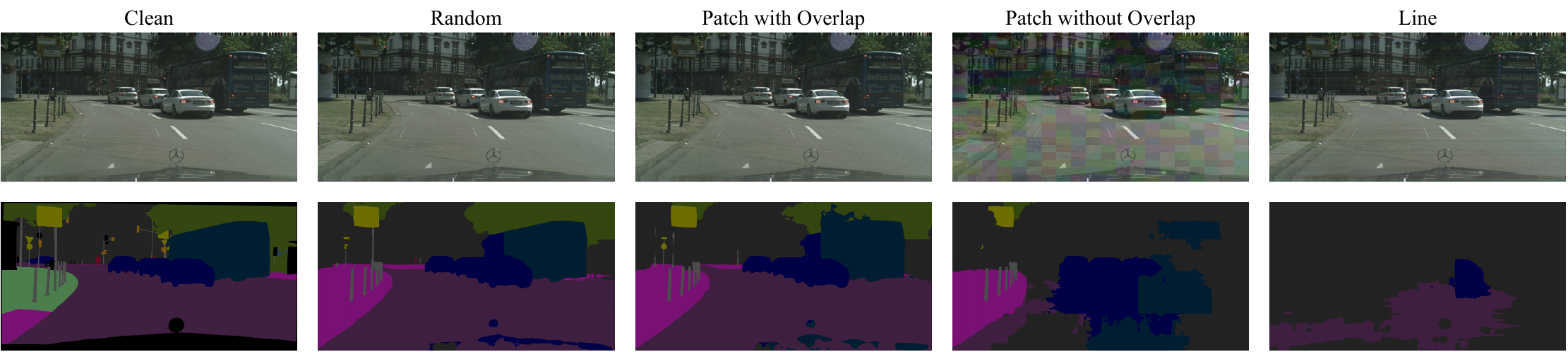}
    \caption{Description of Perturbation Interaction. We use perturbations in the form of random, patch with overlap, patch without overlap, and line to attack, which shows that there is interference between perturbations. Less overlap can lead to better attack performance and linear noises achieve better results in both imperceptibility and attack.}
    \label{fig:interaction}
    \vspace{-8pt}
\end{figure*}

\textbf{Inconsistent Optimization Goal.}
Decision-based attacks on image classification commonly rely on boundary attacks~\cite{qeba}. Boundary attack~\cite{qeba} requires that the image is classified incorrectly, and then minimizes perturbations' magnitude so that the adversarial example is near the decision boundary. However, this strategy cannot be applied in semantic segmentation. As shown in Figure~\ref{fig:eps}, we can add very large noise to make the mean Intersection-over Union (mIoU) very small, but when reducing the noise magnitude, unlike image classification that maintains misclassification, the mIoU also becomes greater, which makes the optimization goal and attack direction inconsistent.

To address this challenge, we propose the utilization of a proxy index to generate adversarial examples from clean images. Our approach involves optimizing the adversarial perturbation starting from clean images and updating the example based on the changes observed in the proxy index associated with the image. To gain a deeper understanding of the proxy index, we propose a simple baseline method called Random Attack. The update process for this baseline method is as follows:
\begin{equation}
    x^0_{adv}=x,\ x^{t+1}_{adv}= \Pi_\epsilon \left(x^t_{adv} + rand[-\frac{\epsilon}{16}, +\frac{\epsilon}{16}] \right),
\end{equation}
where $rand[-\frac{\epsilon}{16}, +\frac{\epsilon}{16}]$ generates a noise that is the same as the input's dimension and satisfies the random distribution $[-\frac{\epsilon}{16}, +\frac{\epsilon}{16}]$, and $\Pi_\epsilon$ clips the input to $[x-\epsilon, x+\epsilon]$. During the iteration, Random Attack updates the perturbation only when the proxy index becomes smaller. The complete algorithm of Random Attack is in \textit{Supplementary Material~B}.

Building upon Random Attack, we conduct a toy study using PSPNet~\cite{pspnet} and SegFormer~\cite{segformer} on Cityscapes~\cite{cityscapes} and ADE20K~\cite{ade20k}, following the same experimental settings as described in Section~\ref{sec:expriments}. Considering that mIoU is a widely adopted metric~\cite{pascal} for evaluating semantic segmentation,  it holds potential as a suitable proxy index. Additionally, the per-pixel classification accuracy (PAcc) can also reflect the attack performance. Hence, we select mIoU and PAcc as the proxy indices, and the attack process using Random Attack is illustrated in Figure~\ref{fig:proxyindex}. Our observations are as follows: \textbf{i)} Random Attack based on the proxy index can reduce the mIoU of the image, \textbf{ii)} when mIoU is employed as the proxy index, the attack performance is superior. This is because when PAcc is used as the proxy index, the adversarial example only needs to maximize misclassification at each pixel, without considering the overall class. Conversely, when mIoU is used as the proxy index, the mIoU of an individual image approaches the mIoU of the entire dataset, resulting in improved attack performance. Therefore, we select mIoU as the proxy index for our study.

\textbf{Perturbation Interaction.}
Despite the effectiveness of Random Attack with the proxy index, as depicted in Figure~\ref{fig:proxyindex}, we observe that its attack performance is constrained and prone to convergence, suggesting that it may have reached a local optimal solution. Recent research~\cite{segpgd} demonstrates that during white-box attacks on semantic segmentation, the classification of each pixel exhibits instability. In one iteration, a pixel may be misclassified, while in the next iteration, it could be classified correctly. This situation also occurs in black-box attacks on semantic segmentation, as shown in Figure~\ref{fig:recover}.

Upon revisiting Random Attack, we hypothesize that there exists interference between the perturbations added in each iteration. This interference arises due to the inconsistent update direction between black-box and white-box attacks. Consequently, a pixel may succeed in one iteration of the attack but fail in the subsequent iteration, leading to convergence towards a local optimal solution. To mitigate this issue, we propose updating the perturbation not on the entire image but on a local region. This localized perturbation update approach may alleviate the interference and enhance the attack performance.

Taking inspiration from this observation, we explore different perturbation update strategies of varying shapes and conduct corresponding experiments. The visualization of these strategies is presented in Figure~\ref{fig:interaction}. When random perturbations are added to the entire image, the resulting segmented mask generally remains close to the original prediction, and the object's outline is relatively well-preserved. This aligns with the limited attack performance depicted in Figure~\ref{fig:proxyindex}. However, when we update the perturbation in the form of patches with overlap~\cite{square}, we observe minimal changes in the attack performance, and the object's outline is still well-maintained. Conversely, when the perturbations are patches without overlap~\cite{bandits}, a significant portion of the object's outline is destroyed, indicating the presence of interference between perturbation updates.

Looking back at the adversarial example in patches without overlap, although its attack effect is significant, it is easy to observe that carefully designed perturbations are added because the added patches are blocky and the color is obvious.
To ensure an effective attack and the perturbation is imperceptible simultaneously, we consider modeling the form of the perturbation as a line, as shown in Figure~\ref{fig:interaction}. We primarily choose linear noises for the following reasons: \textbf{i)} local adversarial perturbations can be spread to the global through context modeling of semantic segmentation~\cite{indirect}, thereby attacking pixels in other areas, so linear noises are still effective. \textbf{ii)} Linear noises are thinner compared to patches, making them relatively harder to detect by the human eye. As depicted in Figure~\ref{fig:interaction}, linear noises exhibit superior performance compared to other strategies while remaining imperceptible.

\begin{table}[t]
\caption{Search strategy for perturbation values. We report the mIoU under 50/200 query budgets and observe that for the same queries, discrete perturbations always obtain lower mIoU (\%).}
\label{tab:pervalue}
\setlength{\tabcolsep}{2mm}
\scalebox{0.75}{
\begin{tabular}{@{}c|c|cccc@{}}
\toprule[1pt]
\multirow{2}{*}{Datset} & \multirow{2}{*}{Model} & \multicolumn{4}{c}{Attack} \\ \cmidrule(l){3-6} 
 &  & Clean & Random & NES & Discrete \\ \midrule
\multirow{2}{*}{Cityscapes} & PSPNet & 77.83 & 48.81/47.18 & 48.34/47.40 & \textbf{33.57/33.54} \\
 & SegFormer & 80.43 & 58.59/56.07 & 58.00/55.34 & \textbf{41.70/41.70} \\ \midrule
\multirow{2}{*}{ADE20K} & PSPNet & 37.68 & 26.63/26.34 & 25.67/25.52 & \textbf{23.50/23.31} \\
 & SegFormer & 43.74 & 34.72/34.45 & 34.66/34.55 & \textbf{33.68/33.53} \\ \bottomrule[1pt]
\end{tabular}
}
\vspace{-8pt}
\end{table}

\textbf{Complex Parameter Space.}
Despite the effectiveness of linear noises in enhancing attack performance, the presence of complex parameter spaces still hampers attack efficiency. Semantic segmentation poses a multi-constraint optimization problem, making it challenging to find the optimal adversarial example within a limited query budget.

In black-box attacks, we usually use two methods to update the perturbation value: random noise~\cite{square,simba} and gradient estimation~\cite{nes,bandits}. Random noise causes clean images to randomly walk on the decision boundary and hope to cross it. Gradient estimation is a gradient-free optimization technology that approximates a gradient direction through random sampling, which can speed up attack efficiency and the most commonly used one is Natural Evolutionary Strategies (NES)~\cite{nes}. 
Although both of the above strategies are effective, they still require many queries, and the query budget increases significantly as the parameter space becomes larger~\cite{bandits}.  
Even if \cite{bandits} introduces prior information to reduce the parameter space, the query efficiency is relatively limited. Therefore, we consider further reducing the parameter space.

For limited queries, it is unlikely to enumerate the entire parameter space. Recent work~\cite{frankwhofe,Parsimonious} shows that adversarial examples are often generated at the extreme points of $l_\infty$ norm ball, which illustrates that it is easier to find adversarial examples at these extreme points. Empirical findings in \cite{Parsimonious} also suggest that adversarial examples obtained from PGD attacks~\cite{pgd} are mostly found on the extreme points of $l_\infty$ norm ball.
Inspired by this, we directly restrict the possible perturbation as the extreme points of the $l_\infty$ norm ball and change the parameter space from continuous space to discrete space. Specifically, the adversarial perturbation $\delta$ is sampled from the Binomial distribution $\{-\epsilon, \epsilon\}^d$, called discrete noises. In this way, we directly reduce the parameter space from $[-\epsilon,\epsilon]^d$ to $\{-\epsilon,\epsilon\}^d$, which only $2^d$ possible search directions.
We conduct a case study to illustrate the effectiveness of these discrete noises, as shown in Table~\ref{tab:pervalue}. Here, we use Random Attack as the baseline and report the mIoU under 50 and 200 query budget. We observe that for the same number of queries, discrete noises can always obtain lower mIoU, and there is a significant gap with other strategies, which illustrates the effectiveness of reducing the parameter space.

\subsection{Discrete Linear Attack}

\begin{algorithm}[t]
  \caption{Discrete Linear Attack (DLA)}
  \label{alg:rla}
  \textbf{Input}: the image $x$, model $f$, proxy index $L$, iteration $T$
  \\
  \textbf{Output}: $x_{adv}$
  \begin{algorithmic}[1]
    \STATE $l_{\min} \leftarrow L(x),\ \hat{\delta} \leftarrow 0,\ i\leftarrow 0,\ M \leftarrow 1,\ n \leftarrow 0$
  \FOR{$t \in [1,\ T]$}
    \IF{$t \leq \frac{T}{5}$}
    \STATE \textit{\textbf{$//$ Perturbation Exploration}}
    \STATE $k \leftarrow t\ \%\  2$
    \STATE $\delta \sim k \cdot \{-\epsilon, \epsilon\}^h + (1-k) \cdot \{-\epsilon, \epsilon\}^w $
        \IF{$l_{\min} > L(x+\delta)$ }
            \STATE $l_{\min} \leftarrow L(x+\delta), \ \hat{\delta} \leftarrow \delta,\ d \leftarrow k$
        \ENDIF
    \ELSE
    \STATE \textit{\textbf{$//$ Perturbation Calibration}}
        \STATE $c \leftarrow d \cdot \left\lceil\frac{h}{2^n}\right\rceil +(1-d)\cdot \left\lceil\frac{w}{2^n}\right\rceil$ 
        \STATE $M[d\times i \times c: d\times (i+1)\times c +(1-d)\times h, (1-d)\times i\times c:(1-d)\times (i+1)\times c+d\times w] \ *= -1$
        \IF{$l_{\min} > L(x+\hat{\delta}\cdot M)$ }
            \STATE $l_{\min} \leftarrow L(x+\hat{\delta}\cdot M), \ \hat{M} \leftarrow M$
        \ELSE
        \STATE $M[d\times i \times c: d\times (i+1)\times c +(1-d)\times h, (1-d)\times i\times c:(1-d)\times (i+1)\times c+d\times w] \ *= -1$
        \ENDIF
             \STATE $i \leftarrow i+1 $
         \IF{ $i==2^n$}
            \STATE $i\leftarrow0,\ n \leftarrow n +1$
        \ENDIF
        \IF{$n==\left\lceil\log_2(d\cdot h+(1-d)\cdot w)\right\rceil+1$}
        \STATE $\hat{\delta} \leftarrow \hat{\delta} \cdot M,\ i \leftarrow 0,\ n \leftarrow 0$
        \ENDIF
    \ENDIF
  \ENDFOR
  \STATE $x_{adv} \leftarrow x+\hat{\delta}\cdot \hat{M}$
    \RETURN $x_{adv}$
  \end{algorithmic}
\end{algorithm}

In this section, we introduce the proposed Discrete Linear Attack (DLA) based on the aforementioned analysis. DLA consists of two main components: perturbation exploration and perturbation calibration.
In the perturbation exploration phase, DLA introduces discrete perturbations in the horizontal or vertical direction to the input, aiming to achieve a better initialization.
In the perturbation calibration phase, DLA dynamically flips the perturbation direction in certain regions based on the proxy index. This allows for iterative updates and calibration of the perturbation. 
The pipeline of DLA is as outlined in Algorithm~\ref{alg:rla}.

\textbf{Perturbation Exploration.} In Section~\ref{sec:analysis}, discrete linear noises can greatly compress the parameter space and improve attack efficiency. Combined with the proxy index and considering the aspect ratio of the image, we initialize the perturbation as follows:
\begin{equation}
    x_{adv} \leftarrow x+\delta,\quad\delta \sim \{-\epsilon, \epsilon\}^d,
\end{equation}
where $d$ denotes the height or weight of images. In perturbation exploration, we alternately sample discrete linear noises with horizontal or vertical directions and add them to the clean image. Then, we calculate the proxy index and retain the adversarial perturbation that obtains the minimum proxy index as $\hat \delta$.

\textbf{Perturbation Calibration.}
Although perturbation exploration has demonstrated high attack performance, the obtained adversarial perturbations still fall short of optimality. This limitation arises from the coarse-grained nature of perturbation exploration, which fails to consider the fine-grained updating of local perturbations. Given the discrete nature of the noise, we propose generating new perturbations by flipping the sign of the existing perturbation.

In the perturbation calibration phase, we adopt a hierarchical approach to randomly flip the sign of local perturbations, thereby further refining the perturbations. This process involves first attempting to flip the global perturbation and subsequently dividing the image into blocks, performing flipping operations on each block. Specifically, we first partition the entire image into blocks, then iterate over each block and flip the sign of the discrete linear perturbation. If the mIoU after flipping is lower, the current perturbation is saved. After traversing the current block, DLA further divides the image into more fine-grained blocks and then traverses. By employing hierarchical blocking and flipping, we aim to obtain the most effective adversarial examples. This operations are outlined in Lines 12-25 of Algorithm~\ref{alg:rla}.

\begin{table*}[t]
\caption{Attack results on Cityscapes and ADE20K. We report mIoU (\%) under 50/200 query budget.}
\vspace{-9pt}
\label{tab:main}
\begin{center}
\setlength{\tabcolsep}{2.8mm}
\scalebox{0.88}{
\begin{tabular}{@{}ccccccc@{}}
\toprule
 &  & \multicolumn{5}{c}{Model} \\ \cmidrule(l){3-7} 
\multirow{-2}{*}{Dataset} & \multirow{-2}{*}{Attack} & FCN~\cite{fcn} & PSPNet~\cite{pspnet} & DeepLab V3~\cite{deeplabv3} & SegFormer~\cite{segformer} & MaskFormer~\cite{maskformer} \\ \midrule
 & Clean & 77.89 & 77.83 & 77.70 & 80.43 & 73.91 \\
 & Random & 35.76/34.94 & 48.81/47.18 & 54.57/52.77 & 58.59/56.07 & 39.09/39.06 \\
 & NES~\cite{nes} & 34.47/33.82 & 48.34/47.40 & 54.32/52.99 & 58.00/55.34 & 51.94/52.56 \\
 & Bandits~\cite{bandits} & 18.17/15.65 & 20.81/17.55 & 29.85/26.73 & 39.43/36.14 & 26.94/26.88 \\
 & ZO-SignSGD~\cite{zosignpgd} & 34.97/34.01 & 46.69/45.80 & 51.83/50.54 & 55.67/54.81 & 49.65/49.59 \\
 & SignHunter~\cite{signhunter} & 23.88/21.67 & 33.52/26.04 & 44.24/35.93 & 41.38/34.18 & 47.05/27.06 \\
 & SimBA~\cite{simba} & 33.74/29.58 & 46.27/40.22 & 54.67/50.17 & 54.17/52.67 & 33.52/32.71 \\
 & Square~\cite{square} & 35.47/35.99 & 48.47/49.18 & 54.45/56.23 & 56.71/52.18 & 50.87/49.84 \\
\multirow{-9}{*}{Cityscapes~\cite{cityscapes}} & Ours & \textbf{3.18/3.07} & \textbf{2.14/2.06} & \textbf{1.79/1.71} & \textbf{18.12/17.78} & \textbf{2.79/2.78} \\ \midrule
 & Clean & 33.54 & 37.68 & 39.36 & 43.74 & {\color[HTML]{1F2328} 45.50} \\
 & Random & 22.85/22.13 & 27.72/27.36 & 25.82/24.81 & 38.02/37.64 & 25.37/24.06 \\
 & NES~\cite{nes} & 24.47/23.96 & 26.57/26.26 & 23.83/23.41 & 36.35/36.06 & 34.78/34.55 \\
 & Bandits~\cite{bandits} & 25.10/23.67 & 25.02/23.93 & 27.52/26.36 & 36.32/35.03 & 26.14/26.91 \\
 & ZO-SignSGD~\cite{zosignpgd} & 23.29/22.94 & 26.82/26.47 & 25.38/24.41 & 35.22/32.18 & 33.32/32.86 \\
 & SignHunter~\cite{signhunter} & 20.15/16.72 & 24.21/20.16 & 25.40/20.48 & 32.56/28.22 & 28.78/16.78 \\
 & SimBA~\cite{simba} & 24.20/21.49 & 26.36/22.92 & 25.56/22.13 & 36.70/34.81 & 35.62/33.18 \\
 & Square~\cite{square} & 23.94/22.90 & 26.87/25.89 & 27.70/26.46 & 35.43/34.76 & 26.29/26.41 \\
\multirow{-9}{*}{ADE20K~\cite{ade20k}} & Ours & \textbf{8.18/7.97} & \textbf{10.19/9.85} & \textbf{11.34/10.67} & \textbf{28.91/27.85} & \textbf{12.14/12.14} \\ \bottomrule
\end{tabular}
}
\end{center}
\vspace{-8pt}
\end{table*}

\section{Experiments}
\label{sec:expriments}

\subsection{Experimental Setup}

\textbf{Datasets.} We attack the semantic segmentation models with two widely used semantic segmentation datasets: Cityscapes~\cite{cityscapes} (19 classes) and ADE20K~\cite{ade20k} (150 classes). Following \cite{indirect} and \cite{segpgd}, we randomly select 150 and 250 images from the validation set of Cityscapes and ADE20K. For evaluation metrics, we choose the standard metric, mean Intersection-over Union (\textbf{mIoU})~\cite{pascal}, a per-pixel metric that directly corresponds to the per-pixel classification formulation. After attacking, the less the mIoU, the better the attack performance.

\noindent\textbf{Models.} We select two types of semantic segmentation models: traditional convolutional models (FCN~\cite{fcn}, DeepLabv3~\cite{deeplabv3}, and PSPNet~\cite{pspnet}), and transformer-based models (SegFormer~\cite{segformer} and MaskFormer~\cite{maskformer}). Please refer to \textit{Supplementary Material C} for more model details.

\noindent\textbf{Attacks.} We select 7 attack algorithms for performance comparison, including zero-order optimization (NES~\cite{nes}, Bandits~\cite{bandits}, ZO-SignSGD~\cite{zosignpgd}, and SignHunter~\cite{signhunter}) and random search (Random attack (Random), SimBA~\cite{simba}, and Square Attack~\cite{square} (Square)). 

\noindent\textbf{Implementation details.}
In all experiments, the maximum perturbation epslion $\epsilon$ is 8. For NES~\cite{nes}, we set the number of queries for a single attack $q = 10$. For Bandit Attack, we set the initial value of patch size $priority_{size} = 20$, and the learning rate $prior_{exploration}=0.1$.
For ZO-SignSGD~\cite{zosignpgd}, we set the same number of single attack queries as NES $q=10$. For Square Attack~\cite{square}, we set the initial value of the fraction of pixels $p_{init}$ is $0.05$. For SimBA~\cite{simba}, we set the magnitude of the perturbation delta as $50$. The setting of SignHunter~\cite{signhunter} is consistent with the original paper. To alleviate the effect of randomness, we report average mIoU (\%) after three attacks.

\begin{figure*}[t]
    \centering
    \includegraphics[width=1\linewidth]{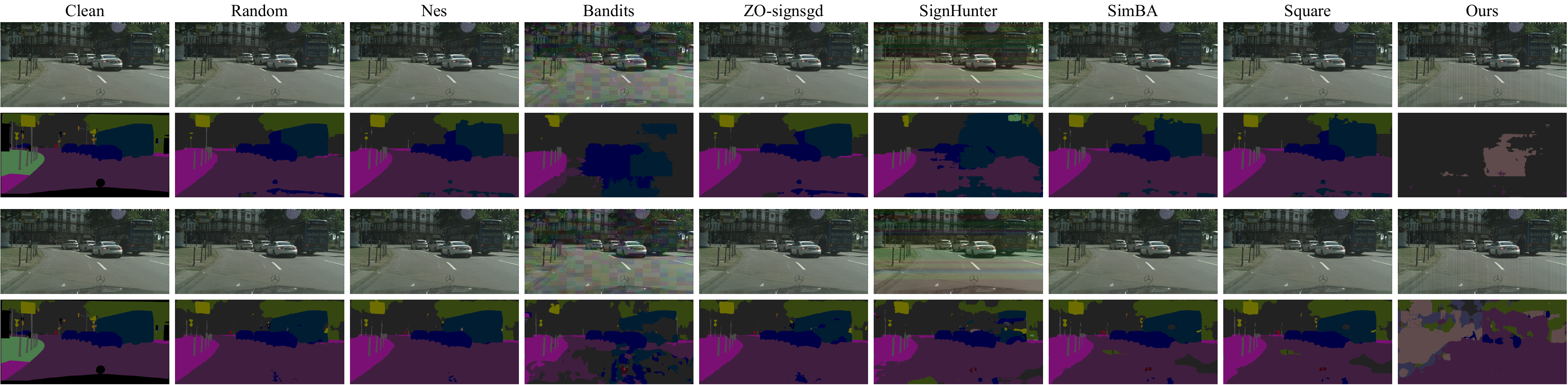}
    \caption{Visualization of different attacks on Cityscapes and the threat model is SegFormer.}
    \label{fig:vis}
    \vspace{-8pt}
\end{figure*}

\subsection{Performance Comparison}
\noindent\textbf{Attack Results.}
Table~\ref{tab:main} illustrates the attack results of 8 black-box attacks on Cityscapes~\cite{cityscapes} and ADE20K~\cite{ade20k}. We report mIoU (\%) of 5 models under 50 and 200 query budget. Random and NES~\cite{nes} have lower attack performance due to their complex parameter spaces. ZO-SignSGD~\cite{zosignpgd}, SimBA~\cite{simba}, and Square~\cite{square} introduce local prior information, which can further improve attack performance. Furthermore, both Bandits~\cite{bandits} and SignHunter~\cite{signhunter} use non-overlapping local noise, thus achieving sub-optimal performance. However, as shown in Figure~\ref{fig:vis}, Bandits' patch noise and SignHunter's strip noise are colored and are very easy to perceive by humans. Our DLA significantly outperforms other competing attacks on both datasets. On Cityscapes' PSPNet, DLA reduces mIoU by 15.49\% and 23.98\% compared to Bandits and Signhunter under 200 queries. Further, on the more challenging PSPNet of ADE20K, DLA reduces mIoU by 14.08\% and 10.31\% compared to Bandits and Signhunter under 200 queries. In terms of visualization, our DLA maintains the imperceptibility of adversarial perturbations and is able to destroy the outline of objects well. In terms of attack efficiency, the attack performance of DLA under 50 queries exceeds the results of other attacks under 200 queries by a very significant gap. Overall, our DLA has extremely high attack efficiency and can more efficiently evaluate the adversarial robustness of existing semantic segmentation models.

\noindent\textbf{Discussion.}
In Table~\ref{tab:main}, we observe that decision-based attacks on ADE20K~\cite{ade20k} are more challenging than attacking Cityscapes~\cite{cityscapes}. We think the possible reason is that the category distribution of images in Cityscapes is relatively even, and they are all urban scenes with relatively high similarity and low complexity, so it is easier to attack. ADE20K has more categories and the differences between images are larger, so the attack is more difficult. In addition, we also find that SegFormer~\cite{segformer} demonstrates the best adversarial robustness under 8 attacks on both datasets, compared with the other 4 semantic segmentation models. This is because SegFormer is a transformer-based model, its main components are transformers, and the self-attention mechanism leads to higher adversarial robustness~\cite{devopatch,shao2022adversarial}, which is consistent with the description in SegFormer. Furthermore, it is worth noting that the backbone of MaskFormer has the structure of CNN, which implies that it does not exhibit a higher level of robustness compared to SegFormer.

\subsection{Diagnostic Experiment}
To study the effect of our core designs, we conduct ablative studies on Cityscapes and ADE20K. We use SegFormer~\cite{segformer} as the threat model and attack it under 50/200 query budget.

\noindent\textbf{Attack Design.}
We first study the attack design of DLA, as shown in Table~\ref{tab:ablation}. In perturbation exploration, \textit{random} is the random noise of Random Attack, and \textit{horizontal} and \textit{vertical} is to add discrete linear noise horizontally and vertically respectively. \textit{iterative} is to add discrete linear noise alternately horizontally and vertically. In perturbation calibration, \textit{random} is the random update noise of Random attack, and \textit{flip} is the update strategy of DLA's filp perturbation sign. We observe that \textit{flip} can achieve better attack performance than \textit{random} in perturbation calibration. When the perturbation exploration is \textit{iterative}, under 200 queries, it exceeds \textit{random} 0.76\% on Cityscapes and 0.83\% on ADE20K. In perturbation exploration, discrete linear noise significantly surpasses random noise by a large advantage. \textit{vertical} and \textit{iterative} achieve the best performance on Cityscapes and ADE20K respectively. We find that the aspect ratio of Cityscapes is fixed, so resulting in vertical noises being more effective. The aspect ratio of ADE20K changes, so \textit{iterative} has better attack performance, which means it is more generalizable when facing images of more scales. Considering the robustness of facing images with different aspect ratios, we choose \textit{iterative} as the strategy for adding discrete linear noises.

\begin{figure}[t]
    \centering
    \includegraphics[width=1\linewidth]{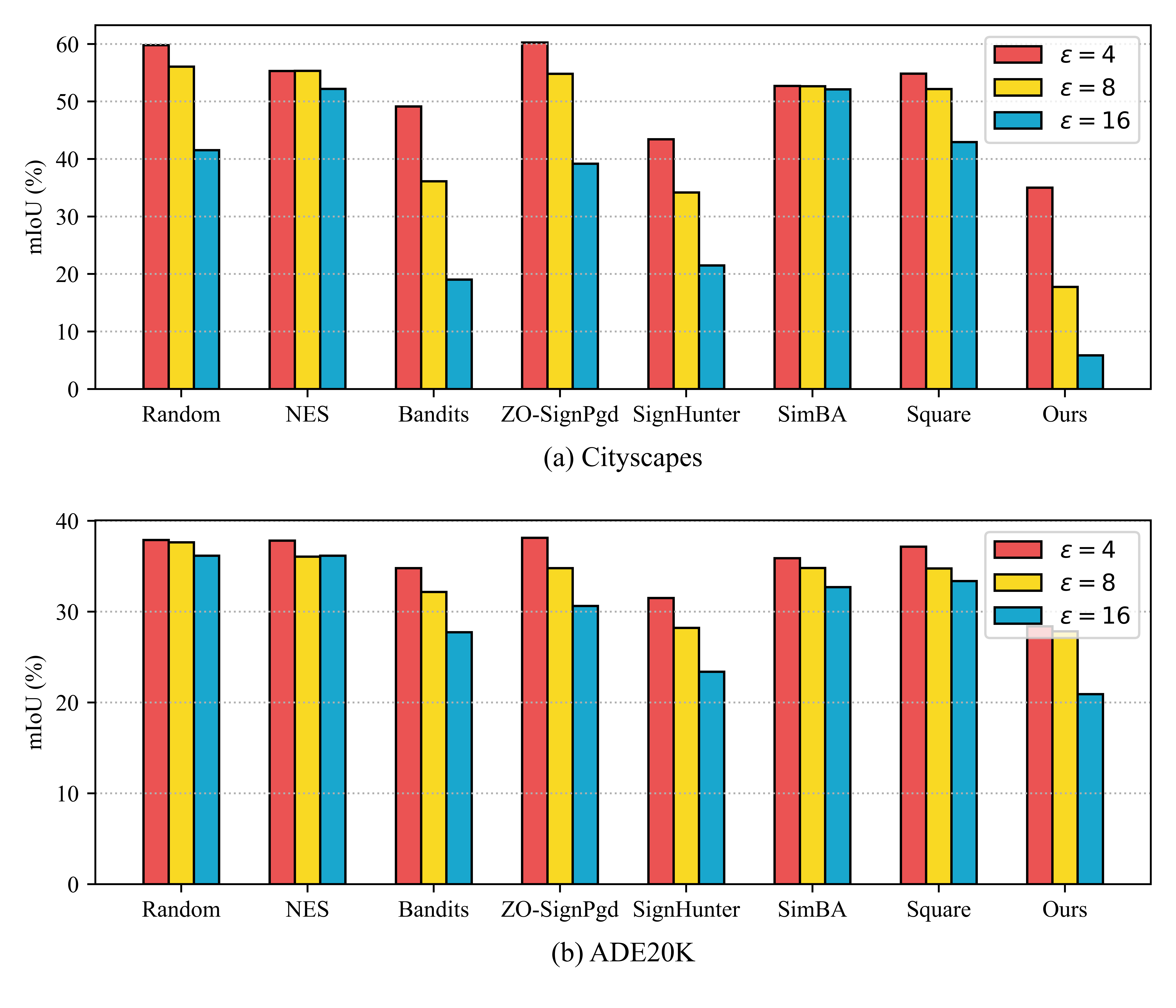}
    \vspace{-16pt}
    \caption{Attack performance of different black-box attacks under different perturbation magnitudes $\epsilon$ within a 200 query budget.}
    \label{fig:model_eps}
    \vspace{-12pt}
\end{figure}

\noindent\textbf{Perturbation Magnitude $\epsilon$.}
To assess the impact of different perturbation magnitudes $\epsilon$ on attack performance, we select $\epsilon$ as 4, 8, and 16 for experiments on SegFormer. Figure~\ref{fig:model_eps} depicts the attack performance of different black-box attacks under different perturbation magnitudes $\epsilon$. As the magnitude of perturbations increases, all attacks exhibit a greater decrease in overall mIoU. Notably, DLA consistently achieves the highest attack performance across all three perturbation magnitudes $\epsilon$. Additionally, we observe that as the magnitude of perturbations increases, DLA outperforms other competing attacks in terms of the extent to which it can degrade mIoU. The above experiments illustrate that DLA has a stronger ability to evaluate the adversarial robustness of semantic segmentation under different perturbation magnitudes than other competing attacks.

\noindent\textbf{Limited Queries.}
Since a large number of queries leads to detection by the target system, we test the attack performance under extremely limited queries. To simulate a limited number of queries, we give 10 query budgets and evaluate the mIoU after the attack, as shown in Table~\ref{tab:limited}. On Cityscapes, DLA demonstrated extreme attack efficiency, reducing SegFormer's mIoU of 61.54\% within 10 queries and surpassing the second-best bandits attack of 22.31\% by a significant margin. In the more challenging ADE20K, our DLA reduces the mIoU of SegFormer by 12.89\% in 10 queries. Likewise, we beat the next best bandits attack by 4.86\%. Combined with the attack results in Table~\ref{tab:pervalue}, our DLA has attack performance that exceeds other current competitive attacks under both limited queries and a large number of queries. It indicates DLA can effectively evaluate the adversarial robustness of semantic segmentation in industrial and academic scenarios.

\begin{table}[t]
\caption{Ablation study on the attack design of DLA.}
\label{tab:ablation}
\setlength{\tabcolsep}{1mm}
\scalebox{0.75}{
\begin{tabular}{@{}cccc|cc|cc@{}}
\toprule
\multicolumn{4}{c|}{Pert. Explor.} & \multicolumn{2}{c|}{Pert. Callbr.} & \multicolumn{2}{c}{Dataset} \\ \midrule
random & horizontal & vertical & iterative & random & flip & Cityscapes & ADE20K \\ \midrule
 &  &  &  &  &  & 80.43 & 43.74 \\
$\checkmark$ &  &  &  & $\checkmark$ &  & 58.59/56.07 & 38.02/37.64 \\
$\checkmark$ &  &  &  &  & $\checkmark$ & 55.47/55.12 & 36.94/36.57 \\
 & $\checkmark$ &  &  &  & $\checkmark$ & 26.41/26.21 & 32.02/31.65 \\
 &  & $\checkmark$ &  &  & $\checkmark$ & \textbf{17.53/17.10} & 29.61/28.49 \\
 &  &  & $\checkmark$ & $\checkmark$ &  & 18.29/18.26 & 29.56/28.77 \\
 &  &  & $\checkmark$ &  & $\checkmark$ & 18.12/17.78 & \textbf{28.91/27.85} \\ \bottomrule
\end{tabular}
}
\end{table}

\begin{table}[t]
\caption{Attack results under limited queries (10 query budget).}
\label{tab:limited}
\centering
\setlength{\tabcolsep}{0.5mm}
\scalebox{0.66}{
\begin{tabular}{@{}clccccccccc@{}}
\toprule
\multicolumn{2}{c}{Attack} & Clean & Random & NES & Bandits & ZO-SignSGD & SignHunter & SimBA & Square & Ours \\ \midrule
\multicolumn{2}{c}{Cityscapes} & 80.43 & 59.22 & 59.26 & 41.20 & 56.23 & 47.90 & 54.62 & 57.22 & \textbf{18.89} \\
\multicolumn{2}{c}{ADE20K} & 43.74 & 37.82 & 38.20 & 35.71 & 37.82 & 35.95 & 37.57 & 37.64 & \textbf{30.85} \\ \bottomrule
\end{tabular}
}
\vspace{-12pt}
\end{table}

\section{Conclusions}
In this paper, we provide the first in-depth study of decision-based attacks on semantic segmentation. For the first time, we analyze the differences between semantic segmentation and image classification and study the three major challenges of corresponding decision-based attacks. Based on random search and proxy index, we discover discrete linear noise and propose a novel discrete linear attack (DLA). We conduct extensive experiments on 2 datasets and 5 models. Compared with 7 attack competitors, DLA has higher attack performance and query efficiency. On Cityscapes, DLA can reduce PSPNet’s mIoU from 77.83\% to 2.14\% within 50 queries. Therefore, DLA is expected to evaluate adversarial robustness in security-sensitive applications.

\noindent\textbf{Broader Impacts.} Our proposed method exhibits exceptional attack performance and efficiency, thereby presenting a significant and concerning threat to the field of semantic segmentation. This threat becomes particularly pronounced when considering its potential deployment in security-sensitive applications, such as medical diagnosis and autonomous driving. Consequently, we anticipate that this threat will serve as a catalyst for the development of robust designs for semantic segmentation models and will also draw increased attention from the public.

{
    \small
    \bibliographystyle{ieeenat_fullname}
    \bibliography{main}
}


\end{document}